\documentclass[a4paper,twoside]{article}
\usepackage{epsfig}
\usepackage{subcaption}
\usepackage{calc}
\usepackage{amssymb}
\usepackage{amstext}
\usepackage{amsmath}
\usepackage{amsthm}
\usepackage{multicol}
\usepackage{pslatex}
\usepackage{apalike}
\usepackage[bottom]{footmisc}

\usepackage{cite}
\usepackage{amsmath,amssymb,amsfonts}
\usepackage{graphicx}
\usepackage{textcomp}
\usepackage{xcolor}
\usepackage{url}

\usepackage[ruled, lined, linesnumbered, commentsnumbered, longend]{algorithm2e}
\DontPrintSemicolon
\usepackage{algpseudocode}

\SetCommentSty{normalfont}
\SetKwInput{KwRequire}{Require}
\SetNlSty{}{}{:}

\usepackage{balance}
\usepackage{subcaption}
\usepackage{caption}
\usepackage{multirow}
\usepackage{booktabs}
\usepackage{array}
\usepackage{arydshln}
\usepackage{dashbox}

\usepackage{enumitem}
\usepackage{balance}

\def\BibTeX{{\rm B\kern-.05em{\sc i\kern-.025em b}\kern-.08em
    T\kern-.1667em\lower.7ex\hbox{E}\kern-.125emX}}

\usepackage{amsmath,amssymb} 
\usepackage{hyperref}
\hypersetup{colorlinks,allcolors=black}

\usepackage{SCITEPRESS}     

\begin{document}

\title{Curriculum for Crowd Counting - Is it Worthy?}

\author{\authorname{Muhammad Asif Khan\sup{1}\orcidAuthor{0000-0003-2925-8841}, Hamid Menouar\sup{1}\orcidAuthor{0000-0002-4854-909X} and Ridha Hamila\sup{2}\orcidAuthor{0000-0002-6920-7371}}
\affiliation{\sup{1}Qatar Mobility Innovations Center, Qatar University, Doha, Qatar}
\affiliation{\sup{2}Department of Electrical Engineering, Qatar University, Doha, Qatar}
\email{mkhan@qu.edu.qa, hamidm@qmic.com, hamila@qu.edu.qa}
}


\abstract{Recent advances in deep learning techniques have achieved remarkable performance in several computer vision problems. A notably intuitive technique called Curriculum Learning (CL) has been introduced recently for training deep learning models. Surprisingly, curriculum learning achieves significantly improved results in some tasks but marginal or no improvement in others. Hence, there is still a debate about its adoption as a standard method to train supervised learning models. In this work, we investigate the impact of curriculum learning in crowd counting using the density estimation method. We performed detailed investigations by conducting ~112 experiments using six different CL settings using eight different crowd models. Our experiments show that curriculum learning improves the model learning performance and shortens the convergence time.}

\onecolumn \maketitle \normalsize \setcounter{footnote}{0}  \vfill

\section{Introduction}
Crowd counting is an interesting problem in computer vision research \cite{khan2022revisiting, Fan_2022, Gouiaa_2021}. Though several methods \cite{Li_2008, Topkaya_2014, Chen_2012, Chan_2009} have been proposed earlier to estimate crowd in an image, the defacto state-of-the-art approach for crowd counting is using density estimation. Density estimation employs a deep learning model such as a convolution neural network (CNN) to estimate the crowd density in an image. The ground truths to train the model are density maps of crowd images. A density map is generated from the dot annotation map where each dot (representing the head position of a person) is convolved with a Gaussian function.
\par
Over the years, several models have been proposed to improve the accuracy performance over benchmark datasets. Notably, CrowdCNN was the first CNN-based model proposed in \cite{CrowdCNN_CVPR2015}. CrowdCNN is a single-column 6-layer CNN network using density map prediction. Due to the single column, CrowdCNN does not capture the scale variations present in head sizes in crowd images. CrowdNet \cite{CrowdNet_CVPR2016} and MCNN \cite{MCNN_CVPR2016} propose multi-column architectures to cope with the scale variations. The CrowdNet model uses a shallow and a deep network to predict different crowd densities. The MCNN model used three columns of CNN layers with different sizes of convolution kernels in each layer to efficiently capture the scale variations. Though, these multi-column networks achieve better accuracy, their performance is poor on highly congested scenes mainly due to two reasons. First, the model's ability to capture scale variations is limited by the number of columns. Second, these shallow models become quickly saturated due to the small number of neurons. To solve the scale variation problem, improved model architectures such as encoder-decoder e.g., TEDnet \cite{TEDnet_CVPR2019}, SASNet \cite{SASNet_AAAI2021}, and pyramid structure using multi-scale modules e.g., MSCNN \cite{MSCNN_ICIP2017}, SANet \cite{SANet_ECCV2018} are proposed. For crowd estimation in highly congested scenes, models using transfer learning from pre-trained models such as VGG-16 \cite{VGG16_ICLR2015}, ResNet \cite{ResNet_CVPR2016}, MobileNet \cite{MobileNetV2_CVPR2018}, and Inception \cite{Inception_CVPR2015} achieved best results.
Few notable models using transfer learning include CSRNet \cite{CSRNet_CVPR2018}, C-CNN \cite{CCNN_ICASSP2020}, BL \cite{BL_ICCV2019} (VGG), MobileCount \cite{MobileCount_PRCV2019} (MobileNet), MMCNN \cite{MMCNN_ACCV2020}, MFCC \cite{MFCC_2022} (ResNet), SGANet \cite{SGANet_IEEEITS2022} (Inception) etc.

Recently, Curriculum Learning (CL) has gained significant attention as an alternative method to improve performance in various deep-learning tasks.
Curriculum learning refers to the set of techniques to train deep learning models by imitating human curricula. In a CL strategy, the training samples are organized in a specific order (typically by increasing or decreasing difficulty) before feeding to the model. CL was first formalized in \cite{CL_ICML2009} inspired by the fact that humans learn better when the tasks are presented in a meaningful order i.e., typically in the order of increasing complexity (or difficulty). CL potentially brings two benefits: (i) faster convergence and (ii) improved accuracy.
\par
CL has been applied in several supervised learning applications including object localization \cite{Ionescu_CVPR2016, Shi_ECCV2016, Tang_2018}, object detection \cite{Chen_ICCV2015, Li_BMVC2017, Sangineto_PAMI2019} and machine translation \cite{Kocmi_RANLP2017, Platanios_NAACL2019, Wang_ACL2019}. CL has also been successfully applied in reinforcement learning \cite{Narvekar_JMLR2020}. Although CL applied in several problems achieved improved training, faster convergence, and performance gains; authors in \cite{Wu_2021} show CL could not benefit the accuracy performance of image classification on CIFAR10 dataset \cite{CIFAR10_dataset2009}. Some recent works also applied CL in crowd density estimation \cite{LCDnet}.

\begin{figure*}
\centering
\includegraphics[width=0.98\textwidth]{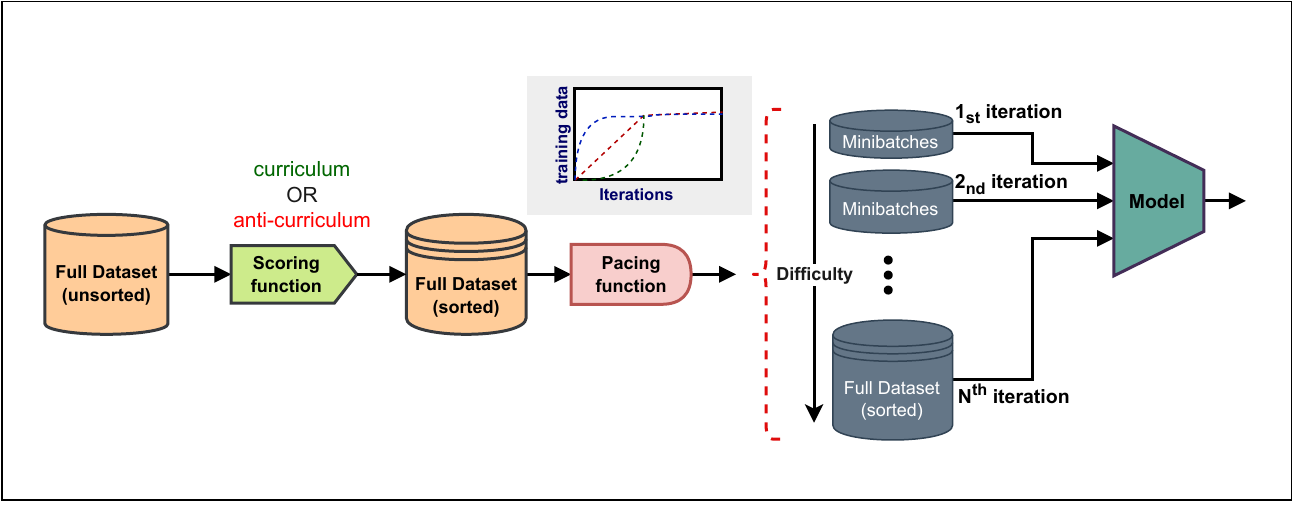}
\caption{Curriculum learning framework.}
\label{fig:CL}
\end{figure*}

This paper aims to perform a rigorous evaluation of CL in yet another important task in computer vision i.e., crowd counting. Crowd counting employs density estimation using pixel-wise regression from crowd images. Recently, very few works have applied CL in crowd density estimation reporting potential benefits in some scenarios. As the results reported in these works are only incremental, this paper aims to extensively investigate the potential of CL in crowd density estimation. The contribution of this paper is as follows:
\begin{itemize}
\item We conducted $\sim 112$ experiments using eight mainstream crowd-counting models and six different CL settings over two benchmark crowd datasets.
\item The models' performance is evaluated using two widely used metrics for crowd counting and results are compared to understand when and to which extent CL outperforms standard learning.
\item Conclusions are drawn for prospective researchers in the area of crowd counting.
\end{itemize}

\section{Background}
Curriculum learning (CL) is defined as "training criteria $\mathcal{C}$ over $T$ training steps: $\mathcal{C} = < Q_1,...Q_t, ... Q_T >$ such that each criterion $Q_t$ is a reweighting of the target training distribution $P(z)$":
\begin{equation} \label{eq:cl}
    Q_t(z) = W_t(z) P(z)  \quad  \forall z \in \text{training \; set} \; D
\end{equation}
In eq: \ref{eq:cl}, (i) the entropy of $P(z)$ gradually increases i.e., $H(Q_t) < H(Q_{t+1})$, (ii) the weight of any example increases i.e., $W_t(z) \leq W_{t+1}(z)$, or (iii) $Q_T(z) = P(z)$ \cite{Wang2022_CL}.

A formal description of the curriculum learning method is presented in Algorithm \ref{algo:cl}.

\begin{algorithm}
\caption{Curriculum Learning} \label{algo:cl}
\KwRequire{pacing function $g$, scoring function $f$, data $X$}
\KwResult{mini-batches [$B_1, B_2, ... B_M$]}

$results$ = sort $X$ using $f$ \;

\For{$i=1, \cdots M$}{
    $size \leftarrow g(i)$ \;
    $X_i = X[1, ..., size]$ \;
    uniformly sample $B_i$ to $results$ \;
    append $B_i$ to $result$ \;
    }
\KwRet{$results$}
\end{algorithm}

  
  
  

There are two main parts of curriculum learning, a scoring function, and a pacing function. The scoring function is used to organize the training samples in a specific meaningful order while the pacing function samples the amount of data exposed to the model in each training step. Fig. \ref{fig:CL} depicts the curriculum learning process.

\subsection{Scoring Function}
A \textit{scoring function} ($f$) is a function that sorts the training data in the order of increasing or decreasing difficulty (in curriculum versus anti-curriculum learning, respectively). For a given scoring function $f: X \rightarrow R$, ($x_i, y_i$) is more difficult than ($x_j, y_j$) if $f(x_,y_i) > f(x_j,y_j)$. A scoring function can be defined in two ways; (i) \textit{self-taught}, or (ii) \textit{transfer-scoring}. In \textit{self-taught} scoring function, the network is trained on uniformly-sampled (randomly ordered) batches to compute the score (difficulty) for each training sample. In \textit{transfer-scoring} function, a pre-trained model is used to compute the score for each training sample.

\subsection{Pacing Function}
\begin{figure}[!h] \centering
\includegraphics[width=0.9\columnwidth]{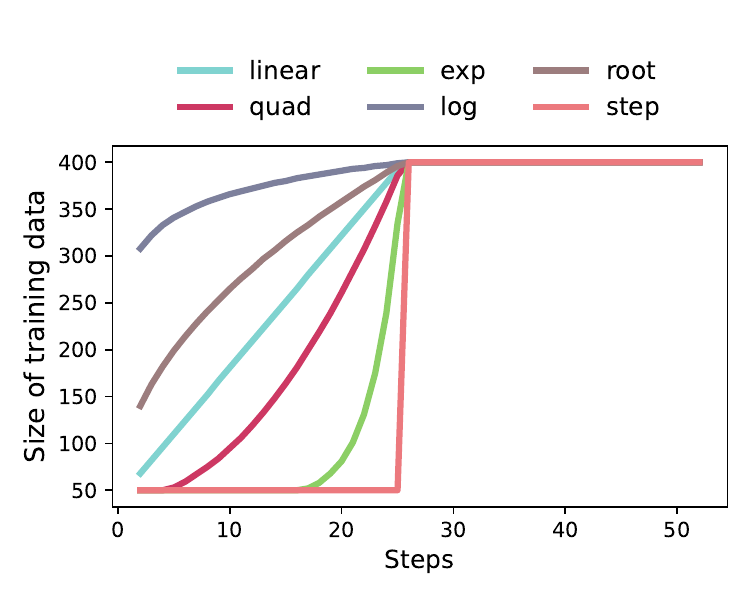}
\caption{Various pacing functions applied to ShanghaiTech Part B dataset with a total number of samples (N) = 400 and batch size = 8.} \label{fig:pacing}
\end{figure}

A \textit{pacing function} $g$ is a function that determines a subset of training data fed to the model in a particular iteration. For training data $X$ of size $N$, $g: [M] \rightarrow [N]$ finds subsets $X_1, X_2, ... X_M \subset X$. From each $X$, mini-batches $\{B_i\}_{i=1}^M$ are uniformly sampled. Fig. \ref{fig:pacing} depicts six different pacing functions used in curriculum learning.

In the more native form of CL, the training examples are organized in order of increasing difficulty. However, some works suggest \textit{anti-curriculum}, in which the training examples are organized in the order of decreasing difficulty.

\section{Related Work} \label{sec:rel_work}
Recently, curriculum learning has been adopted in a few crowd-counting works.
In \cite{Li_2021}, authors propose TutorNet to improve density estimation in crowd counting. A main network that generates density maps for crowd images is supervised at the training stage by a TutorNet network. The network produces a weight map of the same shape as the density map. Each value in the weight map then represents the per-pixel learning rate of the model error. Thus, this weight map is used as a curriculum to train the main network. TutorNet uses ResNet as a frontend to extract features. The work also uses scaling of the density map pixel values. TutorNet is evaluated over ShanghaiTech \cite{MCNN_CVPR2016} and Fudan-ShanghaiTech dataset (FDST) \cite{FDST_dataset}. The results show major improvement using density map scaling with further improvement using TutorNet (pixel-level curriculum). The work did not consider TutorNet alone, hence providing little intuition on the efficacy of CL.
The authors in \cite{Wang_2022} followed a similar approach to implement CL in crowd-counting tasks. A weight is assigned to each pixel in a crowd image which indicates the per-pixel difficulty. More specifically, a region-aware density map (RAD) is first generated through an average pooling operation and then the Gaussian function is applied to RAD to produce an attention map. In the attention maps, the simple pixels are assigned higher weights. A modified loss function is proposed to use the attention maps during the model training. The learning performance is evaluated on ShanghaiTech \cite{MCNN_CVPR2016}, UCF-QNRF \cite{CompositionLoss_2018}, WorlExpo'10 \cite{CrowdCNN_CVPR2015}, and GCC \cite{GCC_dataset} datasets.
\par
The computation of pixel-wise curricula can be a more expensive task as compared to sample-wise curricula in many works on CL. A recent study \cite{khan2023clip} on curriculum learning integrated with dataset pruning to improve the learning performance and convergence time supports the efficacy of sample-wise curriculum learning.
\par
A review of the aforementioned works provides limited hints on the efficacy of curriculum learning in crowd-counting tasks. However, whether the performance gains in these works are the results of the curriculum learning, the underlying crowd model, or of both together? Whether curriculum learning can improve the performance of any kind of crowd models e.g., shallow, deep, multi-column, encoder-decoder, and multi-scale, with and without transfer learning? This work aims to further investigate to answer these questions.

\section{Experiments and Evaluation} \label{sec:exp}
We conducted more than 112 experiments using eight (8) mainstream crowd models and two well-known datasets to investigate the efficacy of curriculum learning in crowd counting. 

\subsection{Datasets}
The two datasets chosen are ShanghaiTech Part A and ShanghaiTech Part B datasets, both published in \cite{MCNN_CVPR2016}. The datasets contain cross-scene crowd images with varying crowd densities and have been extensively used for benchmarking in numerous studies on crowd counting and density estimation.

\subsection{Baseline Crowd Models}
We choose eight (8) different crowd models to use in our experiments. These include MCNN \cite{MCNN_CVPR2016}, CMTL \cite{CMTL_AVSS2017}, MSCNN \cite{MSCNN_ICIP2017}, CSRNet \cite{CSRNet_CVPR2018}, SANet \cite{SANet_ECCV2018}, TEDnet \cite{TEDnet_CVPR2019}, Yang et al. \cite{Yang2020}, and SASNet \cite{SASNet_AAAI2021}. The eight crowd models are chosen such that they vary in terms of model size, complexity, and design. 

\subsection{Curriculum Settings}
We consider six different types of pacing functions: linear, quadratic, exponential, root, logarithmic, and step. These pacing functions are calculated using Eq. \ref{eq:pacing}.

\begin{equation} \label{eq:pacing}
g = 
\begin{cases}
Nb + aNb  & (linear) \\[0.5em]
Nb + N \frac{1-b}{aT} t^P - p=1/2, 1, 2   &(quad) \\[0.5em]
Nb + \frac{N \left(1-b\right)}{e^{10}-1} \left( exp\left( \frac{10t}{aT} - 1 \right)\right) &(exp) \\[0.5em]
Nb + N \left(1-b\right) \left(1+\frac{1}{10}log\left(\frac{t}{aT} + e^{-10}\right)\right) &(log) \\[0.5em]
Nb + N \left[ \frac{x}{aT} \right]  &(step) \\[0.5em]
\end{cases} 
\end{equation}

For each dataset, we kept the value of $b$ fixed, while choosing the value of $a = [0.2, 0.4, 0.6, 0.8]$. Thus, in each experiment, we take a fraction ($b$) of full training data and incrementally add batches according to the pacing function (with parameter $a$).
The pacing functions used are plotted in Fig. \ref{fig:pacing}.

\begin{figure*}
\centering
\subcaptionbox{Linear pacing function} {\includegraphics[width=0.31\textwidth]{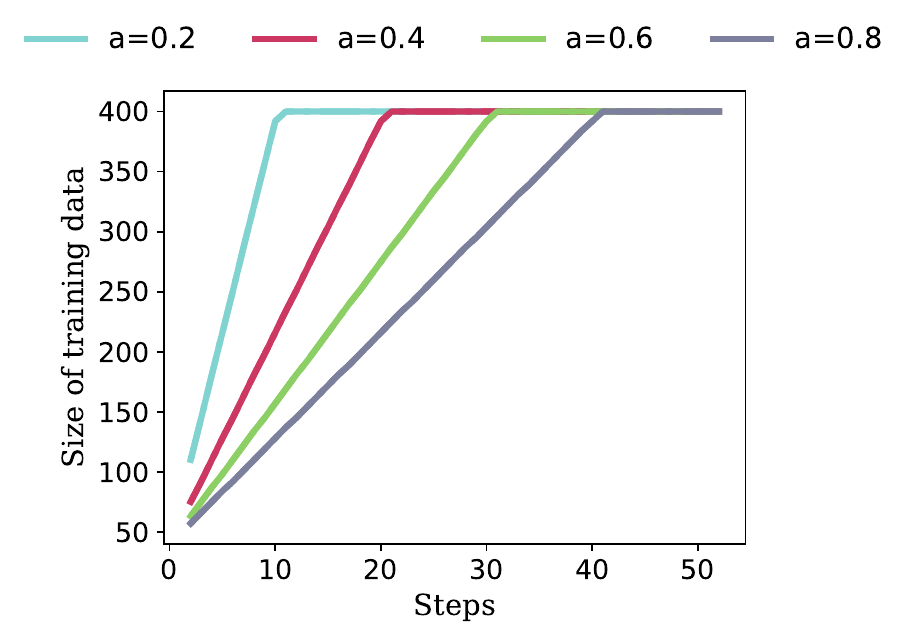}} \hspace{1em}
\subcaptionbox{Quadratic pacing function} {\includegraphics[width=0.31\textwidth]{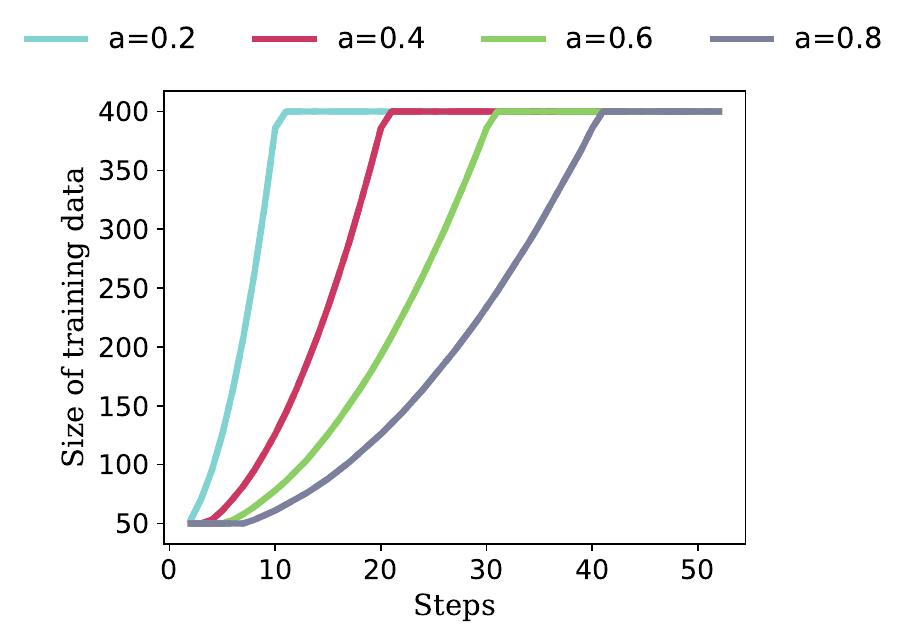}} \hspace{1em}
\subcaptionbox{Exponential pacing function} {\includegraphics[width=0.31\textwidth]{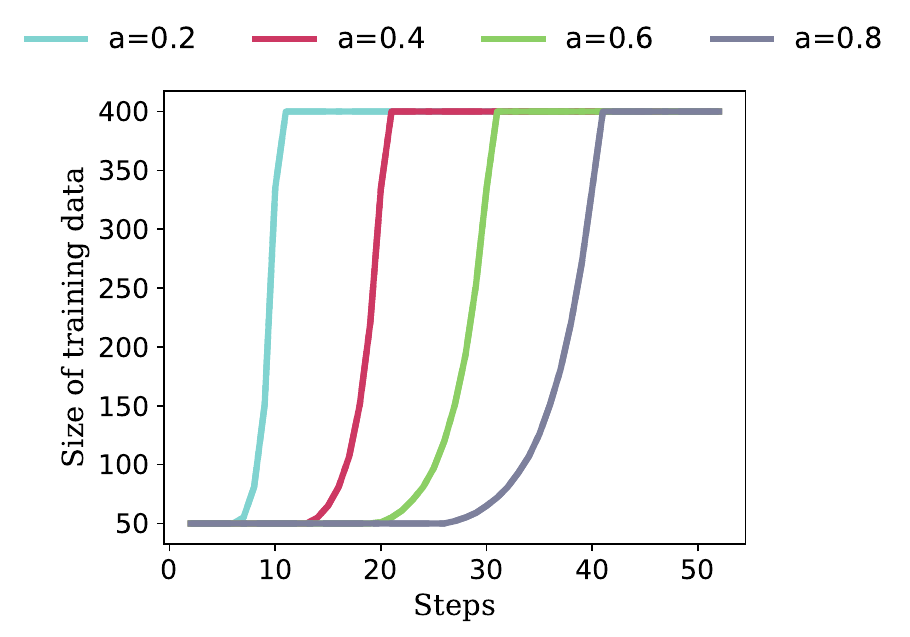}} \\
\vspace{2em}
\subcaptionbox{Root pacing function} {\includegraphics[width=0.31\textwidth]{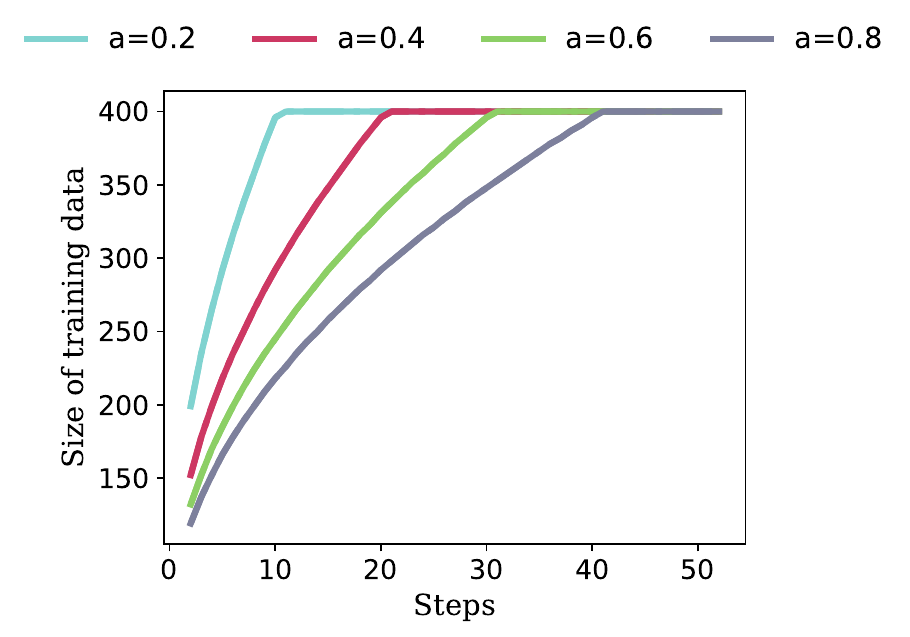}} \hspace{1em}
\subcaptionbox{Log pacing function} {\includegraphics[width=0.31\textwidth]{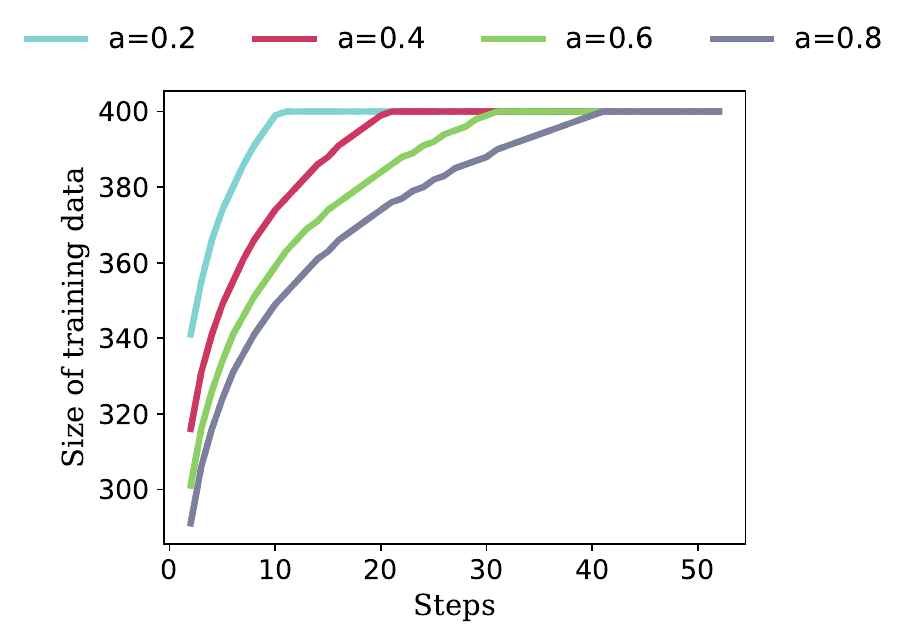}} \hspace{1em}
\subcaptionbox{Step pacing function} {\includegraphics[width=0.31\textwidth]{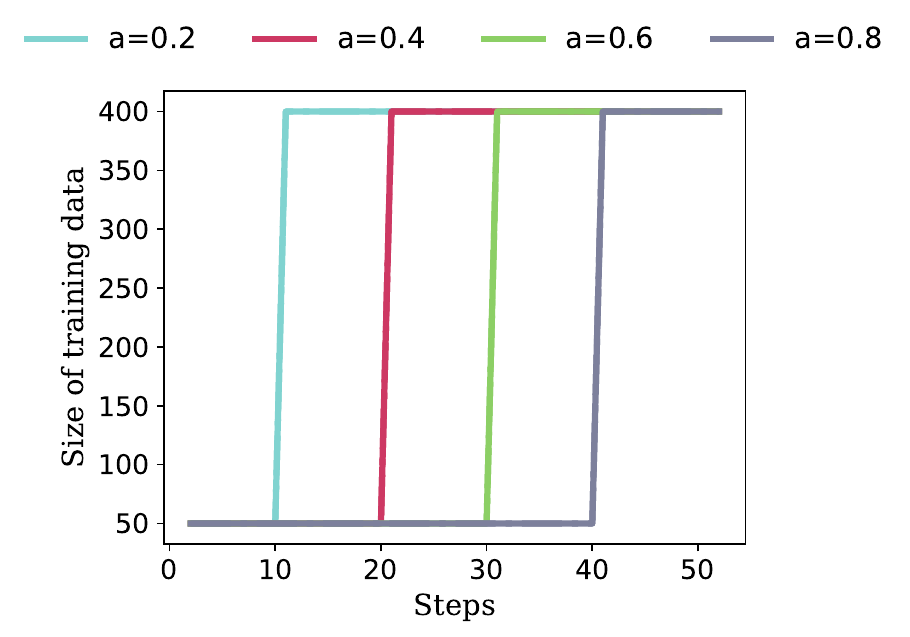}} \hspace{1em}
\caption{Pacing functions used in our experiments.}
\label{fig:pacing_functions}
\end{figure*}

\subsection{Evaluation Metrics}
We used two commonly used metrics to test the performance in the crowd-counting task i.e., mean absolute error (MAE) and mean squared error (MSE).
MAE and MSE can be calculated using Eq. \ref{eq:mae} and Eq. \ref{eq:mse}, respectively.

\begin{equation} \label{eq:mae}
    MAE = \frac{1}{N} \sum_{1}^{N}{(e_n - g_n)}
\end{equation}

\begin{equation} \label{eq:mse}
    MSE = \frac{1}{N} \sum_{1}^{N}{(e_n - g_n)}
\end{equation}

where $N$ represents the total number of examples in the dataset, $g_n$ is the actual count of people in the $n^{th}$ crowd image, and ${e_n}$ is the estimated count (computed as the sum of pixel values in the predicted density map for the same image). Other less frequently used metrics are grid average mean error (GAME) for more localized counting, structural similarity index (SSIM), and peak signal-to-noise ratio (PSNR) for the quality of the predicted density maps.

\subsection{Training Details}
Each crowd model is trained first on the ShanghaiTech Part B dataset using standard training. The same model is trained (from scratch) using curriculum learning with a single pacing function. Since there are six pacing functions used in this study, the same model is trained six times with a different pacing function. In each experiment, we carefully selected the pacing function parameter $\alpha$ to define reasonable subsets in each iteration by examining the relative performance over a few epochs (Fig. \ref{fig:pacing_functions}). 
Thus, we conduct a total of seven (7) complete training for a single model and an overall 56 experiments on the ShanghaiTech Part B dataset. The same number of experiments are then repeated for the ShanghaiTech Part A dataset. All models are trained using the PyTorch framework on two RTX-8000 GPUs. In all experiments, we use Adam optimizer with an initial learning rate of $1\times10^{-2}$ and a \textit{ReduceLROnPlateau} learning rate decay function based on MAE values.

\section{Results and Analysis} \label{sec:results}
The trained model in each experiment is evaluated over the two metrics (i.e., MAE and MSE). We carefully chose the values of parameters $a$ and $b$ in the pacing functions to achieve reasonable subsets of training data in curriculum learning settings. In standard training, the batches are uniformly sampled from the full training dataset. The best-achieved results in each experiment are depicted in Table \ref{tab_STB} and \ref{tab_STA}.

\begin{table*}[!h]
\centering
\caption{A comparison of standard training versus curriculum learning (using six different pacing functions) over ShanghaiTech Part B dataset using two metrics (MAE and MSE). The bold text shows the lowest error values.}
\label{tab_STB}
\small

\begin{tabular}{r| cc| cc| cc| cc| cc| cc| cc} \toprule[0.15em]

\multirow{3}{*}{Model} 
&\multicolumn{2}{c|}{Standard} &\multicolumn{12}{c}{Curriculum Learning} \\ \cmidrule{2-15}

&\multicolumn{2}{c|}{Random} &\multicolumn{2}{c|}{Linear} &\multicolumn{2}{c|}{Log} &\multicolumn{2}{c|}{Quadratic} &\multicolumn{2}{c|}{Exponential} &\multicolumn{2}{c|}{Step} &\multicolumn{2}{c}{Root} \\ \cmidrule{2-15}
&MAE &MSE  &MAE &MSE  &MAE &MSE  &MAE &MSE  &MAE &MSE  &MAE &MSE  &MAE &MSE \\ \midrule \midrule

MCNN        & 26.4 & 41.3 & \textbf{19.2} & \textbf{32.2} & 23.8 & 38   & 21.4 & 33.8 & 23.1 & 37.4 & 23.4 & 37.4 & 22.8 & 37.1 \\

CMTL         & 20.0   & 31.1 & 19.6 & 30.6 & 19.8 & 30.7 & \textbf{18.8} & \textbf{30.4} & 20.0   & 31.6 & 20.2 & 32.0   & 19.4 & 30.5 \\

MSCNN        & 17.7 & 30.2 & 16.9 & 29.2 & 17.6 & 29.9 & 17.2 & 29.4 & 17.8 & 30.2 & 17.8 & 30.1 & \textbf{16.8} & \textbf{28.8} \\

CSRNet       & 10.6 & 16.0   & \textbf{7.8}  & \textbf{14.2} & 10.2 & 16   & 8.2  & 14.6 & 9.4  & 15.3 & 9.8  & 15.8 & 8.6  & 14.9 \\

SANet       & 8.4  & 13.6 & 8.4  & 13.5 & 8.8  & 14.0   & \textbf{8.1}  & \textbf{13.3} & 8.5  & 13.8 & 8.2  & 13.4 & 8.6  & 13.7 \\

TEDnet      & 8.2  & 12.8 & \textbf{7.6}  & \textbf{12.2} & 8.2  & 12.7 & 8.3  & 13   & 7.7  & 12.4 & 8.1  & 12.6 & 7.8  & 12.4 \\

Yang et al. & 12.3 & 21.2 & 11.8 & 20.4 & 11.4 & 20.0   & \textbf{10.6} & \textbf{18.3} & 12.2 & 21.4 & 12.6 & 21.6 & 12.0   & 20.8 \\

SASNet      & 6.4  & 9.9  & 6.6  & 10   & \textbf{6.3}  & 9.6  & \textbf{6.8}  & 10.5 & 7.2  & 11.2 & 6.9  & 10.6 & 6.4  & 9.8 \\
\bottomrule
\end{tabular}
\end{table*}

We note several interesting observations in the results. First, curriculum learning clearly brings significant improvements in some cases. For instance, on the ShanghaiTech Part B dataset, the MAE for MCNN was reduced from 26.4 to 19.2 using the linear pacing function (best case) and 21.4 using the quadratic pacing function (second best case). Similarly, the MAE for CSRNet was reduced from 10.6 to 7.8 using the linear pacing function. On the ShanghaiTech Part A dataset, the MAE for MCNN was reduced from 110.2 to 102.4 (linear pacing function), and for CSRNet the MAE was reduced from 68.2 to 58.4.

Second, curriculum learning brings marginal improvement in most cases. This is evident from the MAE values of all models for several selections of pacing functions.
Third, curriculum learning could not improve or underperforms the standard training in some cases (indicated with red font color in both tables). This observation highlights the importance of the pacing function as an important hyperparameter in curriculum learning.
Fourth, the benefit of curriculum learning is evident for all models on both datasets. However, the level of improvement varies among the models which raise the logical question of what makes curriculum learning outperform standard training.
The last observation is that the benefits of each pacing function have been consistent to a certain level. For instance, the best results were produced by linear function for MCNN \cite{MCNN_CVPR2016}, CSRNet \cite{CSRNet_CVPR2018}, and TEDnet \cite{TEDnet_CVPR2019} on both datasets. Similarly, SASNet \cite{SASNet_AAAI2021} produces better results using the log pacing function on both datasets. The Step function almost underperformed all other pacing functions except for SANet \cite{SANet_ECCV2018} and TEDnet \cite{TEDnet_CVPR2019}.
\par

\begin{table*}[htbp]
\centering
\caption{A comparison of standard training versus curriculum learning (using six different pacing functions) over ShanghaiTech Part A dataset using two metrics (MAE and MSE). The bold text shows the lowest error values.}
\label{tab_STA}
\small

\begin{tabular}{r| cc| cc| cc| cc| cc| cc| cc} \toprule[0.15em]

\multirow{3}{*}{Model} 
&\multicolumn{2}{c|}{Standard} &\multicolumn{12}{c}{Curriculum Learning} \\ \cmidrule{2-15}

&\multicolumn{2}{c|}{Random} &\multicolumn{2}{c|}{Linear} &\multicolumn{2}{c|}{Log} &\multicolumn{2}{c|}{Quadratic} &\multicolumn{2}{c|}{Exponential} &\multicolumn{2}{c|}{Step} &\multicolumn{2}{c}{Root} \\ \cmidrule{2-15}
&MAE &MSE  &MAE &MSE  &MAE &MSE  &MAE &MSE  &MAE &MSE  &MAE &MSE  &MAE &MSE \\ \midrule \midrule

MCNN  & 110.2 & 173.2 & \textbf{102.3} & \textbf{150.8} & 109.6 & 172.4 & 108.8 & 170.6 & 106.7 & 169.0 & 112.4 & 177.6 & 102.4 & 151.2 \\ 

CMTL  & 101.3 & 152.4 & \textbf{96.7} & \textbf{142.4} & 100.6 & 151.8 & 98.2 & 149.2 & 103.5 & 156.2 & 105.4 & 160.3 & 100.2 & 150.4 \\ 

MSCNN  & 83.8 & 127.4 & 82.3 & 122.6 & 83.4 & 128.3 & 82.6 & 125.8 & 83.8 & 129.2 & 85.8 & 133.5 & \textbf{81.2} & \textbf{120.8} \\ 

CSRNet & 68.2 & 115.0 & \textbf{58.4} & \textbf{102.5} & 62.0 & 105.4 & 60.6 & 105.2 & 62.2 & 106.1 & 62.0 & 105.5 & 62.3 & 106.4 \\ 

SANet  & 67.0 & 104.5 & 66.9 & 103.8 & 68.4 & 105.4 & \textbf{66.2} & \textbf{100.8} & 66.4 & 101.2 & 70.2 & 118.4 & 67.2 & 105.0 \\

TEDnet  & 64.2 & 109.1 & \textbf{60.2} & \textbf{102.9} & 65.8 & 114.2 & 66.4 & 115.2 & 65.0 & 111.3 & 63.4 & 106.4 & 63.8 & 106.9 \\

Yang et al. & 104.6 & 145.2 & 102.3 & 139.0 & 104.4 & 143.7 & \textbf{98.4} & \textbf{134.2} & 104.8 & 145.8 & 105.6 & 147.2 & 99.1 & 134.8 \\

SASNet  & 53.6 & 88.4 & \textbf{51.2} & \textbf{81.0} & 51.4 & 81.4 & 52.8 & 86.2 & 53.2 & 87.0 & 55.8 & 90.4 & 52.3 & 84.6 \\
\bottomrule
\end{tabular}
\end{table*}

\begin{figure}
    \centering
    \includegraphics[width=0.9\columnwidth]{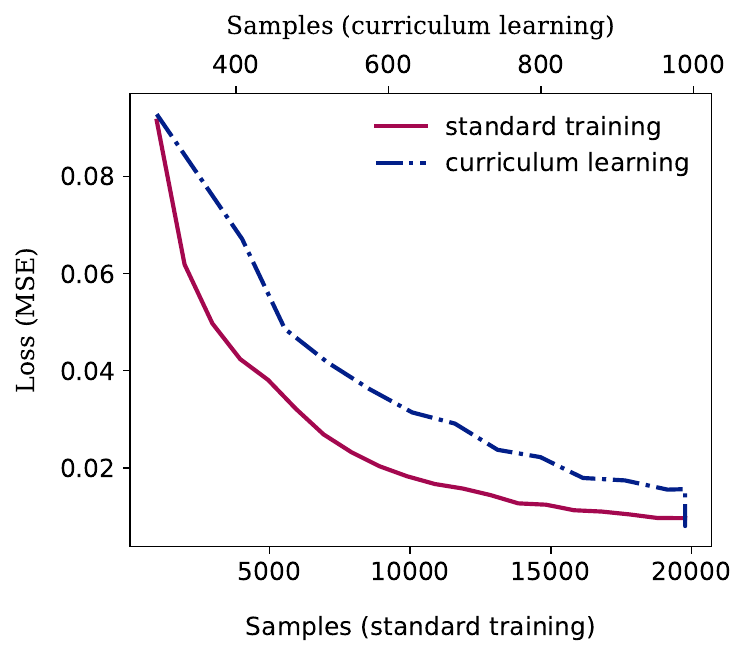}
    \caption{Illustration of model convergence using standard training versus curriculum learning.}
    \label{fig:convergence}
\end{figure}

Besides the potentially significant results discussed previously, Fig. \ref{fig:convergence} depicts the clear benefit of curriculum learning in terms of convergence time. The y-axis shows the MSE loss of the model during the training phase, whereas the top and bottom x-axes show the number of samples seen by the model during the curriculum learning and standard training settings. The loss drops too quickly in curriculum learning as compared to standard training highlighting the faster convergence of curriculum learning.

\section{Conclusions} \label{sec:conclusion}
This article presents a detailed experimental analysis of curriculum learning in crowd-counting. Although curriculum learning has been effective in reinforcement learning, its efficacy in supervised learning is not fully evident due to the lack of detailed investigations. We performed an extensive set of experiments on eight mainstream crowd-counting models to evaluate the performance of curriculum learning. The results show significant improvements in some cases, marginal improvements in most cases with no improvement in a few cases. Through the detailed analysis of the results, we conclude that curriculum learning can potentially improve the performance of deep learning models by carefully choosing the pacing function and its parameters. Moreover, given the short training time budget, curriculum learning is a good choice to cut the convergence time. For future work, we suggest extending the investigation to other computer vision tasks using large datasets and different scoring functions appropriate to the task.

\section*{Acknowledgement}
This publication was made possible by the PDRA award PDRA7-0606-21012 from the Qatar National Research Fund (a member of The Qatar Foundation). The statements made herein are solely the responsibility of the authors.

\bibliographystyle{apalike}
{\small
\bibliography{manuscript}}

\end{document}